\documentclass[letterpaper, 10 pt, conference]{ieeeconf}  

\usepackage{graphicx}
\usepackage{caption}
\usepackage{amsmath}
\usepackage{amssymb}
\usepackage{booktabs}
\usepackage{multirow}
\usepackage{wrapfig}
\IEEEoverridecommandlockouts                              

\title{\LARGE \bf
Learning Holistic Whole-Body Loco-Manipulation with a Bipedal Mobile Manipulator
}
\author{Zhongyu Chen$^{1,*}$, Yuxuan Nai$^{1,*}$, Qian Chen$^{1}$, Yidong Zhu$^{1}$, Chen Jing$^{1}$,\\
Qihan Wang$^{1}$, Xudong Li$^{1}$, Zhizhan Li$^{1}$, Leixin Chang$^{1}$, Liangjing Yang$^{1}$, Hua Chen$^{2,\dagger}$%
\thanks{$^{*}$Equal contribution. $^{\dagger}$Corresponding author.}%
\thanks{$^{1}$ZJU-UIUC Institute. $^{2}$LimX Dynamics.}%
}

\let\oldtwocolumn\twocolumn
\renewcommand\twocolumn[1][]{%
  \oldtwocolumn[{#1%
    \begin{center}
      \includegraphics[width=0.92\textwidth]{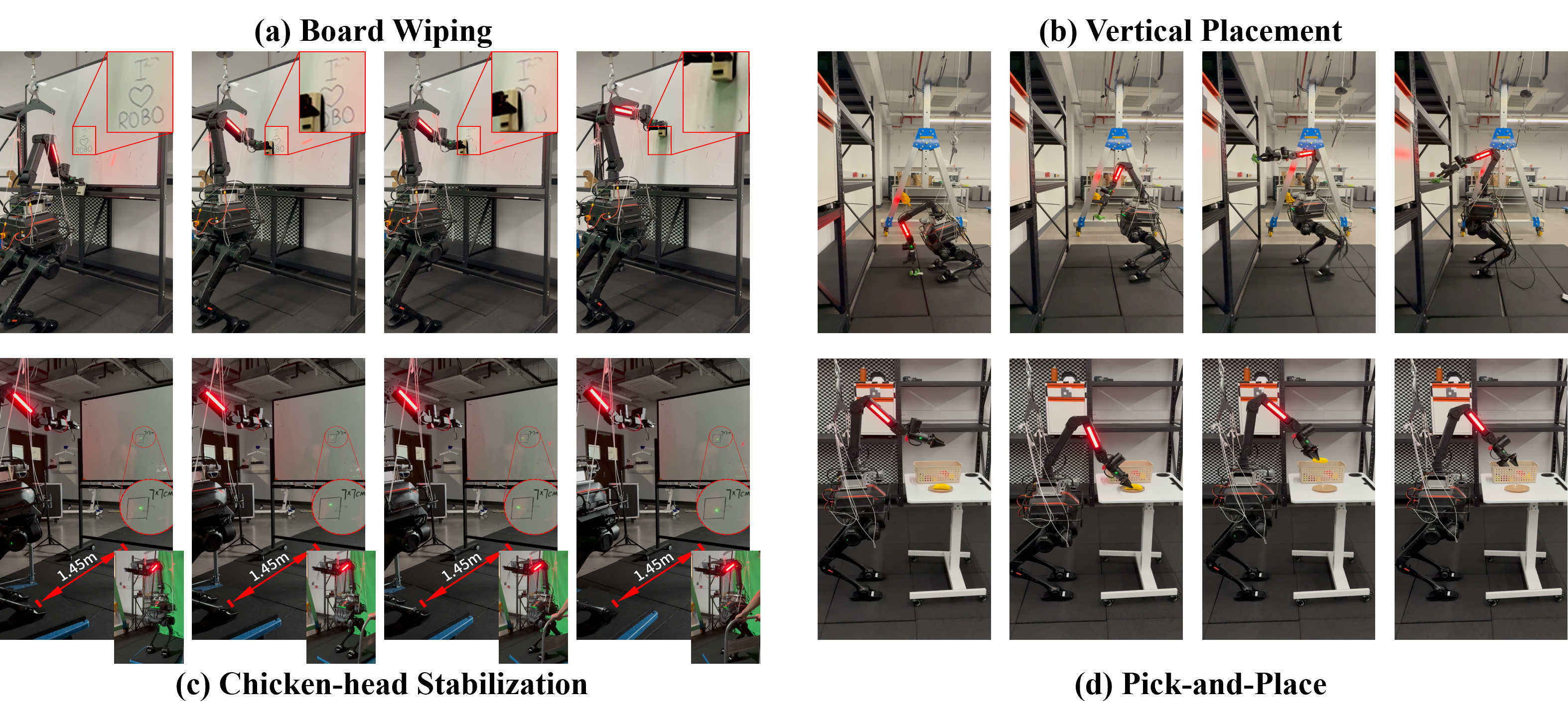}

      \vspace{2pt}
      \captionof{figure}{
      Overview of our controller. Given only a 6-DoF end-effector target,
      the learned whole-body controller coordinates arm and leg motions
      to produce diverse whole-body manipulation behaviors.
      }
      \label{fig:overview}
      \vspace{4pt}
    \end{center}
  }]%
}

\begin{document}

\maketitle
\thispagestyle{empty}
\pagestyle{empty}

\begin{abstract}
Bipedal loco-manipulation enables robots to interact with objects beyond the nominal workspace of their arms by coordinating locomotion and manipulation. Realizing this capability requires a low-level whole-body controller that translates task-level manipulation goals into coordinated arm and leg motions while maintaining balance. We present a unified whole-body controller trained with reinforcement learning that directly maps 6-DoF end-effector targets to coordinated actions for the bipedal base and robotic arm. Given only an end-effector target, the learned controller autonomously coordinates reaching, postural adaptation, and stepping without explicit base-velocity or footstep commands. A reward-gating strategy regulates the trade-offs among end-effector tracking, locomotion, and balance during training, while a temporal context estimator combines windowed Transformer encoding, recurrent GRU memory, and auxiliary dynamics prediction to extract dynamics-relevant information from observation history. Real-robot experiments demonstrate that the same controller supports reaching, postural adaptation, and stepping under commands from VR teleoperation, a learned diffusion policy, and scripted trajectories, providing a common end-effector interface for diverse manipulation tasks.

\end{abstract}

\section{Introduction} \label{sec:introduction}

Manipulation tasks often involve reaching objects at different heights
and locations, including targets beyond the nominal workspace of a
fixed-base arm. Loco-manipulation extends the robot's reach by coordinating
arm motion with body posture adjustment and locomotion~\cite{fu2023deep,ha2025umionlegs}.
Effective coordination allows the
robot to use its mobility in service of manipulation, reaching targets
that would otherwise remain inaccessible to the arm alone.

We adopt a bipedal mobile manipulator that combines a two-legged base
with a single six-joint robotic arm. Its low hardware cost and fourteen
controlled joints make it a minimalist platform for studying single-arm
loco-manipulation without requiring a full humanoid upper body.
Despite this structural simplicity, coordinating manipulation with
bipedal mobility remains challenging: arm motion and postural adjustment
must be coordinated with stepping while maintaining dynamic balance.
This motivates a whole-body controller that uses the available arm
and leg motions to accomplish manipulation goals.

For reuse across manipulation tasks, this controller should accept
task-level goals and determine the required whole-body motion.
The motion needed to reach an end-effector target depends on the
robot's current configuration: arm motion may suffice for one target,
whereas another may require postural adjustment or stepping.
Requiring human operators or high-level policies to specify these
body motions ties manipulation commands to the details of whole-body
coordination. An end-effector pose provides a geometrically interpretable
command that specifies the manipulation objective while leaving
whole-body motion generation to the controller. This separation allows
different high-level command sources to share a common execution
interface.

Learning-based whole-body control has enabled coordinated locomotion
and manipulation on legged platforms~\cite{fu2023deep,liu2025visualwbc,portela2025whole}.
UMI-on-Legs connects manipulation policies to a quadrupedal controller
through end-effector trajectories~\cite{ha2025umionlegs}.
Jiang et al. demonstrate direct 6-DoF end-effector pose tracking on
a wheeled-quadrupedal manipulator using a nonlinear reward fusion
module (RFM)~\cite{jiang2025rfm}. For humanoids, ULC jointly tracks
root, torso, and arm commands~\cite{sun2025ulc}, while CEER provides
an end-effector--root interface for heterogeneous high-level planners~\cite{luo2026ceer}.
Building on these advances, we study whole-body
control of a bipedal single-arm platform using a single instantaneous
6-DoF end-effector target as the sole task-level command. The controller
must determine the required postural adjustment and stepping without
additional locomotion or body-posture commands.

This formulation presents two coupled challenges. First, end-effector
tracking, locomotion, and balance impose competing requirements on the
same system. The controller must learn when to recruit the lower body
to support reaching without sacrificing stability to reduce tracking
error. Second, deployment under partial observations requires the
controller to infer dynamics-relevant information from observation
history. Coordinated whole-body responses therefore depend on both
the regulation of training objectives and the representation of
temporal information.

To address these challenges, we present a controller trained with reinforcement learning for end-effector-driven whole-body control of a bipedal
manipulator. Building on RFM~\cite{jiang2025rfm}, we introduce phase-specific
safety scores with lower-bounded gating and a best-so-far progress
reward to coordinate locomotion, balance, and end-effector tracking. We further develop a
temporal context estimator that combines windowed Transformer encoding,
recurrent GRU memory, and auxiliary dynamics prediction to extract
dynamics-relevant information from proprioceptive history.
The resulting policy uses deployable observations and a single
end-effector target to generate coordinated arm and leg actions,
including reaching, postural adaptation, and stepping.

\begin{figure*}[!t]
    \centering
    \includegraphics[width=1\textwidth]{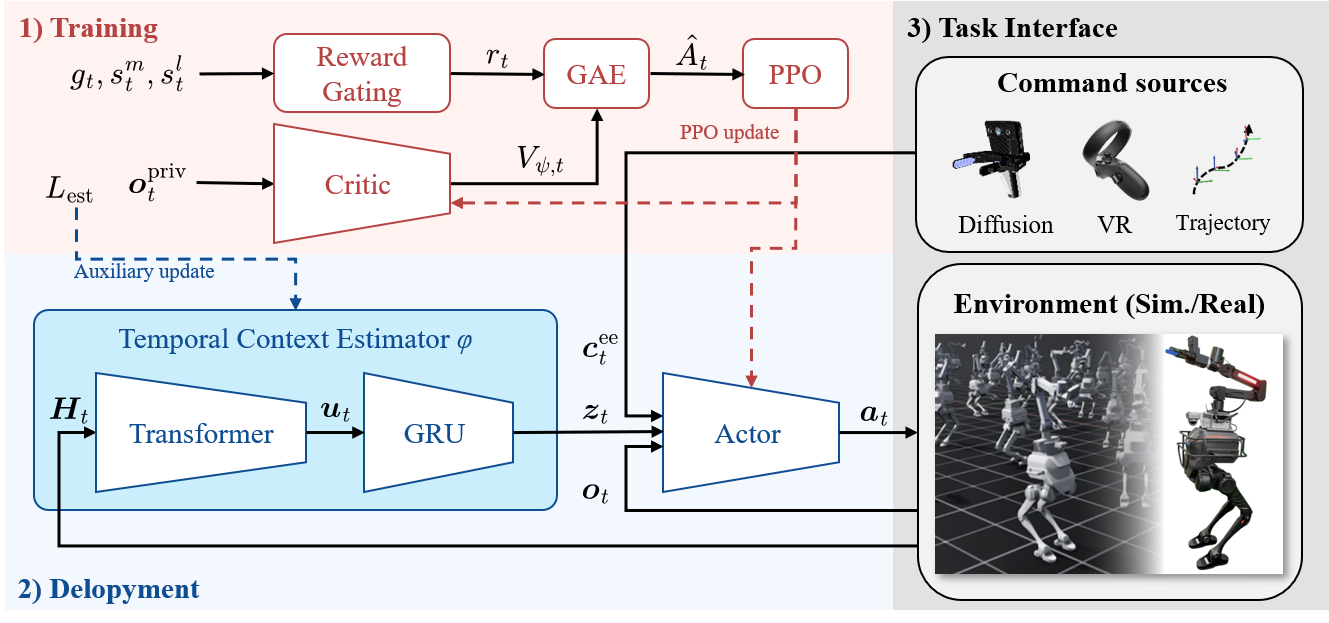}
    \caption{Overview of the proposed controller.
    Given a 6-DoF end-effector target and proprioceptive observations,
    the policy coordinates whole-body actions through reward-gated
    training and temporal context estimation.}
    \label{fig:framework}
\end{figure*}

We integrate our controller into a real bipedal manipulation system with onboard
policy execution, state estimation, and a common end-effector command
representation. Commands from VR teleoperation, a learned diffusion
policy, and scripted trajectories are transformed into this shared
representation and executed by the same whole-body policy.
Simulation comparisons and ablations evaluate task success, tracking
accuracy, and action smoothness. Real-robot experiments demonstrate
manipulation across these command sources and evaluate vertical
end-effector reachability against a floating-base inverse-kinematics
controller. These evaluations examine both the learning components
and the controller's use as a shared manipulation execution layer.
Our main contributions are threefold:
\begin{itemize}

\item We formulate and realize end-effector-driven whole-body control
for a bipedal manipulator, using a single 6-DoF target as the sole
task-level command to coordinate arm motion, postural adaptation,
and stepping without explicit locomotion or body-posture commands.

\item We develop a unified learning framework for bipedal whole-body
manipulation that combines phase-aware reward gating with a
Transformer--GRU temporal context estimator. Simulation ablations
demonstrate that both components contribute to end-effector tracking
and whole-body stability.

\item We integrate and deploy the controller on a low-cost bipedal
manipulation system, demonstrating reuse across VR teleoperation,
a learned diffusion policy, and scripted trajectories, together with
expanded vertical reachability relative to a floating-base
inverse-kinematics baseline.

\end{itemize}

\section{Related Work}

\subsection{Learning-Based Whole-Body Loco-Manipulation}

Learning-based whole-body control coordinates locomotion and
manipulation on legged platforms. Fu et al.~\cite{fu2023deep}
demonstrate unified arm-leg control on a quadrupedal
manipulator, extending end-effector reach through arm-body coordination.
Liu et al.~\cite{liu2025visualwbc} further investigate visual
whole-body control for legged manipulation.
Humanoid approaches organize coordination through unified,
modular, or hierarchical control. ULC~\cite{sun2025ulc} learns
a single whole-body tracking policy, whereas
SEEC~\cite{jang2025seec} learns upper-body compensation for
locomotion-induced disturbances and supports different lower-body
controllers. HiWET~\cite{cao2026hiwet} separates high-level
world-frame reasoning from low-level dynamic execution.

In contrast to prior work centered on quadrupedal manipulators or full humanoids, we study a complementary point in this design space: holistic loco-manipulation on a minimalist bipedal single-arm platform that expands task-relevant end-effector reachability through coordinated postural adaptation and stepping, where a single policy jointly controls all 14 arm and leg joints, rather than decomposing upper- and lower-body control, and resolves the resulting coordination problem under bipedal balance constraints.

\subsection{Task-Space Interfaces for Loco-Manipulation}

Task-space interfaces determine how motion decisions are divided between high-level command sources and low-level controllers. Portela et al.~\cite{portela2025whole} learn 6-DoF end-effector pose tracking but handle locomotion with a separate policy. UMI-on-Legs~\cite{ha2025umionlegs} instead connects manipulation policies to a quadrupedal whole-body controller through task-frame end-effector trajectories.
Other approaches retain explicit body-motion commands. ULC~\cite{sun2025ulc} tracks root, torso, and arm commands, while CEER~\cite{luo2026ceer} provides an end-effector--root interface for high-level planners.

Most closely related, Jiang et al.~\cite{jiang2025rfm} directly track 6-DoF end-effector poses on a wheeled-quadrupedal manipulator without floating-base commands. Our reward-gating formulation builds on their RFM in the bipedal single-arm setting.

Against this landscape, our controller couples an end-effector-only task interface with holistic whole-body control. Given a single instantaneous 6-DoF end-effector target, the policy coordinates arm motion, postural adaptation, and stepping without explicit base or gait commands. The same controller can therefore execute targets generated by VR teleoperation, learned manipulation policies, and scripted trajectories.

\section{Method}
\label{sec:method}

Our goal is to learn a whole-body controller that enables a
bipedal manipulator to track a desired end-effector pose while
maintaining balance. Given the target pose and robot observations,
the controller outputs joint-position commands for both the arm
and legs. We combine temporal context estimation with reward-gated
training to account for partially observed dynamics and balance
the competing demands of tracking and whole-body motion.

\subsection{Training Pipeline}
\label{sec:training_pipeline}
\label{sec:policy_formulation}

Figure~\ref{fig:framework} illustrates the complete training
pipeline. At each control step, the desired end-effector pose is
expressed in the instantaneous base frame. A temporal estimator
encodes recent observations using a Transformer and carries context
across successive windows using a GRU. It produces a base
linear-velocity estimate and a dynamics latent, which form the
context $\mathbf{z}_t$. The actor combines this context with the
current observation $\mathbf{o}_t$ and end-effector command
$\mathbf{c}^{\mathrm{ee}}_t$ to generate whole-body actions:
\begin{equation}
\begin{aligned}
    \mathbf{c}^{\mathrm{ee}}_t
    &=
    \left[
        (\mathbf{p}^{*}_{BE,t})^\top,\,
        \rho_6(\mathbf{R}^{*}_{BE,t})^\top
    \right]^\top,\\
    \mathbf{a}_t
    &\sim
    \pi_\theta
    \left(
        \cdot \mid
        \mathbf{o}_t,
        \mathbf{c}^{\mathrm{ee}}_t,
        \mathbf{z}_t
    \right),\\
    \mathbf{q}^{\mathrm{des}}_t
    &=
    \mathbf{q}^{\mathrm{default}}
    + \mathbf{s}\odot\mathbf{a}_t.
\end{aligned}
\label{eq:ee_command}
\end{equation}
Here, $\mathbf{p}^{*}_{BE,t}$ and $\mathbf{R}^{*}_{BE,t}$ are
the desired end-effector position and orientation in the base
frame. The mapping $\rho_6$ concatenates the first two columns
of the rotation matrix to obtain a continuous 6D
representation~\cite{zhou2019continuity}, giving
$\mathbf{c}^{\mathrm{ee}}_t\in\mathbb{R}^{9}$.
Table~\ref{tab:policy_inputs} lists the 58-dimensional observation
and the context inputs. The task-progress reference
$d_t^{\mathrm{ref}}$ is generated internally. The end-effector pose
is the sole external task command; no reference base velocity,
gait phase, contact schedule, footstep command, or discrete
behavior mode is provided.

The normalized action
$\mathbf{a}_t\in\mathbb{R}^{14}$ contains position offsets for eight
lower-body joints and six arm joints. Here, $\odot$ denotes
element-wise multiplication and $\mathbf{s}\in\mathbb{R}^{14}$
contains joint-specific action scales. These scales are $0.6$ for
the lower body, $0.3$ for arm
joints J1--J3, and $0.2$ for J4--J6. Low-level joint-position
controllers track the resulting targets.

A scheduled SE(3)-distance reference defines a continuous
locomotion--manipulation coefficient. The reward combines this
coefficient with phase-specific safety scores to coordinate target
approach, precise tracking, and whole-body stability. Lower-bounded
gating preserves the learning signal during recovery, while a
best-so-far progress term rewards further reductions in tracking
error. Section~\ref{sec:reward_modulation} gives the complete
formulation.

During training, we sample 6-DoF end-effector targets throughout the
robot's whole-body workspace, including targets reachable through arm
motion alone and targets that require postural adaptation or stepping.
We apply domain randomization to the friction coefficients, base mass,
initial base state, and initial leg-joint configuration. We additionally
apply periodic velocity perturbations and uniform noise to the
proprioceptive and end-effector observations.

The resulting rollouts are used to update an asymmetric actor--critic
with PPO~\cite{pinto2018asymmetric,schulman2017ppo}. The actor uses
only the deployable inputs listed in Table~\ref{tab:policy_inputs}.
The critic additionally receives base linear velocity, joint torques
and accelerations, base height, end-effector velocities, foot positions,
control gains, base mass, and contact parameters for value estimation.
The temporal estimator is optimized through the auxiliary objectives
defined in Sec.~\ref{sec:temporal_estimation}. During deployment,
only the actor and temporal estimator are retained.

\begin{table}[t]
    \centering
    \caption{Deployable inputs to the controller.}
    \label{tab:policy_inputs}
    \setlength{\tabcolsep}{5pt}
    \begin{tabular}{lcc}
        \toprule
        Input term & Symbol & Dimension \\
        \midrule
        \multicolumn{3}{c}{\emph{Robot observation $\mathbf{o}_t$}} \\
        \midrule
        Base angular velocity
        & $\boldsymbol{\omega}^{B}_t$ & 3 \\
        Projected gravity
        & $\mathbf{g}^{B}_t$ & 3 \\
        Joint positions
        & $\mathbf{q}^{\mathrm{obs}}_t$ & 14 \\
        Joint velocities
        & $\dot{\mathbf{q}}_t$ & 14 \\
        Previous action
        & $\mathbf{a}_{t-1}$ & 14 \\
        End-effector pose
        & $\mathbf{x}^{\mathrm{ee}}_t$ & 9 \\
        Task-progress reference
        & $d^{\mathrm{ref}}_t$ & 1 \\
        \midrule
        \multicolumn{3}{c}{\emph{Task command}} \\
        \midrule
        End-effector target
        & $\mathbf{c}^{\mathrm{ee}}_t$ & 9 \\
        \midrule
        \multicolumn{3}{c}{\emph{Temporal context}} \\
        \midrule
        Estimated base velocity
        & $\hat{\mathbf{v}}_t$ & 3 \\
        Dynamics latent
        & $\boldsymbol{\xi}_t$ & 64 \\
        \bottomrule
    \end{tabular}
\end{table}

\subsection{Temporal Context Estimation}
\label{sec:temporal_estimation}

Variations in robot and environment dynamics alter the motion
response produced by a given control action. Although these
variations are not directly observable, their effects appear in
the temporal evolution of the robot state. We therefore infer a
dynamics-relevant context from recent onboard observations and
use it to condition the whole-body control policy.

At control step $t$, the estimator receives the most recent
$H=10$ observations:
\begin{equation}
    \mathcal{H}_t =
    \left[
        \mathbf{h}_{t-H+1},
        \ldots,
        \mathbf{h}_t
    \right]^\top
    \in\mathbb{R}^{H\times 57}.
    \label{eq:observation_history}
\end{equation}
Each $\mathbf{h}_t\in\mathbb{R}^{57}$ contains base angular
velocity, projected gravity, joint positions and velocities,
estimated joint torques, and the measured end-effector pose.
All components are available from onboard sensing or state
estimation at deployment.
We encode this history with a
Transformer~\cite{vaswani2017attention}:
\begin{equation}
    \mathbf{u}_t =
    f_{\mathrm{TF},\phi_{\mathrm{TF}}}
    \left(\mathcal{H}_t\right).
    \label{eq:transformer_history_encoder}
\end{equation}
Positional encodings preserve temporal order, while
self-attention models relationships among robot responses at
different steps within the window. A GRU~\cite{cho2014learning}
then propagates information across consecutive windows:
\begin{equation}
    \left(
        \mathbf{r}_t,\mathbf{m}_t
    \right)
    =
    f_{\mathrm{GRU},\phi_{\mathrm{GRU}}}
    \left(
        \mathbf{u}_t,\mathbf{m}_{t-1}
    \right),
    \label{eq:recurrent_context}
\end{equation}
where $\mathbf{m}_t$ is the recurrent state and $\mathbf{r}_t$
is the corresponding GRU output.

The estimator combines the GRU output with the Transformer
representation through the residual connection implemented in
the network:
\begin{equation}
    \mathbf{e}_t =
    \mathbf{u}_t+
    f_{\mathrm{out}}(\mathbf{r}_t).
    \label{eq:context_residual}
\end{equation}
We partition the resulting estimator output as
\begin{equation}
    \mathbf{e}_t =
    \left[
        \hat{\mathbf{v}}_t^\top,\,
        \boldsymbol{\mu}_t^\top,\,
        \left(\log\boldsymbol{\sigma}_t^2\right)^\top
    \right]^\top
    \in\mathbb{R}^{131},
    \label{eq:context_partition}
\end{equation}
where $\hat{\mathbf{v}}_t\in\mathbb{R}^{3}$ estimates the
local base linear velocity, and
$\boldsymbol{\mu}_t,
\log\boldsymbol{\sigma}_t^2\in\mathbb{R}^{64}$
parameterize a diagonal Gaussian distribution.
Using the reparameterization trick~\cite{kingma2014autoencoding},
we then sample the latent and construct the policy context:
\begin{equation}
\begin{aligned}
    \boldsymbol{\sigma}_t
    &=
    \sqrt{\exp(\log\boldsymbol{\sigma}_t^2)+10^{-4}},\\
    \boldsymbol{\xi}_t
    &=
    \boldsymbol{\mu}_t+
    \boldsymbol{\sigma}_t\odot\boldsymbol{\epsilon}_t,
    \quad
    \boldsymbol{\epsilon}_t
    \sim\mathcal{N}(\mathbf{0},\mathbf{I}),\\
    \mathbf{z}_t
    &=
    \left[
        \hat{\mathbf{v}}_t^\top,\,
        \boldsymbol{\xi}_t^\top
    \right]^\top
    \in\mathbb{R}^{67}.
\end{aligned}
\label{eq:estimated_context}
\end{equation}
The context $\mathbf{z}_t$ is supplied to the actor together
with the current observation and end-effector command, as
defined in Sec.~\ref{sec:training_pipeline}.

During training, an MLP decoder predicts the next-step target
observation from the sampled latent:
\begin{equation}
    \hat{\mathbf{y}}_{t+1}
    =
    f_{\mathrm{dec}}(\boldsymbol{\xi}_t).
    \label{eq:next_observation_prediction}
\end{equation}
The target $\mathbf{y}_{t+1}\in\mathbb{R}^{49}$ is the
configured next-observation vector. It contains base angular
velocity, projected gravity, joint positions and velocities,
end-effector pose, and foot positions. It is distinct from the
actor observation $\mathbf{o}_{t+1}$.
Together with velocity supervision and latent regularization,
this prediction defines the estimator objective:
\begin{equation}
\begin{aligned}
    \mathcal{L}_{\mathrm{est}}
    ={}&
    \left\|
        \hat{\mathbf{v}}_t-\mathbf{v}_t
    \right\|_1
    +
    \left\|
        \hat{\mathbf{y}}_{t+1}-\mathbf{y}_{t+1}
    \right\|_1\\
    &+
    \beta D_{\mathrm{KL}}
    \left[
        \mathcal{N}
        \left(
            \boldsymbol{\mu}_t,
            \operatorname{diag}(\boldsymbol{\sigma}_t^2)
        \right)
        \,\Vert\,
        \mathcal{N}(\mathbf{0},\mathbf{I})
    \right].
\end{aligned}
\label{eq:estimator_loss}
\end{equation}
We set the KL weight to $\beta=0.1$.
The context is detached during PPO updates, so the estimator receives
gradients only from these auxiliary objectives.
The prediction objectives encourage the estimator to encode
the robot's recent motion response, which provides the actor
with information relevant to dynamics adaptation. The recurrent
state is reset at the beginning of each episode.

\subsection{Biped-Aware Reward Gating}
\label{sec:reward_modulation}

Let $e_t^p$ and $e_t^R$ denote the end-effector position
and geodesic orientation errors.
The reference and phase variable follow
RFM~\cite{jiang2025rfm}:
\begin{equation}
\begin{aligned}
    d_0^{\mathrm{ref}} &=2e_0^p+e_0^R,\\
    d_t^{\mathrm{ref}}
    &=\max\left(d_{t-1}^{\mathrm{ref}}-v^{\mathrm{ref}}\Delta t,\,0\right),\\
    g_t &=\sigma\left(k(d_t^{\mathrm{ref}}-\mu)\right).
\end{aligned}
\label{eq:rfm_phase}
\end{equation}
Here, $\sigma$ is the sigmoid function,
$v^{\mathrm{ref}}$ sets the reference decay rate,
and $k$ and $\mu$ control the transition.
The phase variable $g_t$ is computed from a scheduled
SE(3)-distance reference that decreases over the command duration.
A large $g_t$ emphasizes locomotion and whole-body target-approach
rewards, whereas a small $g_t$ emphasizes precise end-effector
tracking and whole-body stabilization.

Locomotion and stationary manipulation admit different safe
body motions. The locomotion score considers foot geometry,
base height, and base orientation, allowing the motion needed
for stepping. The manipulation score additionally considers
base, end-effector, and joint velocities, as well as arm
configuration, to suppress residual motion near the target.
For each phase $q\in\{m,l\}$, we define
\begin{equation}
\begin{aligned}
    E_t^q &=\frac{1}{Z_q}\sum_{j\in\mathcal{C}_q}
    \alpha_j^q\frac{\delta_{j,t}}{\tau_j},\\
    s_t^q &=\exp(-E_t^q/\sigma_s^2),\\
    \tilde{s}_t^q &=s_t^q+c_s,\qquad c_s=0.4.
\end{aligned}
\label{eq:phase_safety_error}
\end{equation}
Here, $\mathcal{C}_q$ denotes the corresponding set of safety
quantities, with $m$ denoting manipulation and $l$ locomotion.
Each deviation $\delta_{j,t}$ is normalized by tolerance
$\tau_j$, weighted by $\alpha_j^q$, and aggregated using
normalization constant $Z_q$.
The parameter $\sigma_s$ controls the decay of the
safety score. A severe transient disturbance can drive $s_t^q$
close to zero. Directly using this score as a multiplicative gate
may suppress the task-learning signal needed for recovery.
The offset $c_s$ relaxes the gate to retain this signal.
The relaxed score $\tilde{s}_t^q$ gates task rewards, while
$s_t^q$ enters the explicit safety reward.

RFM uses a cumulative-error term to regulate task progress.
For bipedal target approach, we instead reward reductions
relative to the best position and orientation errors already
achieved for the current command. This formulation provides
a positive signal only when the controller makes further
tracking progress. Defining
$\bar e_t^p=\min_{0\leq j\leq t}e_j^p$
and
$\bar e_t^R=\min_{0\leq j\leq t}e_j^R$,
we use
\begin{equation}
\begin{aligned}
    r_t^{\mathrm{prog}}
    ={}&2[\bar e_{t-1}^p-e_t^p]_+
    \exp(-\bar e_t^p/\sigma_{\mathrm{prog}})\\
    &+[\bar e_{t-1}^R-e_t^R]_+
    \exp(-\bar e_t^R/\sigma_{\mathrm{prog}}),
\end{aligned}
\label{eq:progress_reward}
\end{equation}
where $[x]_+=\max(x,0)$ and
$\sigma_{\mathrm{prog}}$ sets the error scale.
We retain RFM's position--orientation prioritization and
reference-tracking design~\cite{jiang2025rfm}. The position reward
uses micro-enhancement to increase sensitivity near the target. A
coarse orientation term acts throughout the command, while a fine
orientation term is activated by small position error and the
manipulation gate. Defining
$e_t^{\mathrm{ref}}=[|d_t^{\mathrm{ref}}-d_t|-\gamma]_+$,
the individual terms and their scales are summarized in
Table~\ref{tab:reward_terms}. The resulting task reward is
\begin{equation}
\begin{aligned}
    r_t^{\mathrm{task}}
    ={}&w_p(1-g_t)\tilde{s}_t^m r_t^p
    +w_{R,c}r_t^{R,c}\\
    &+w_{R,f}(1-g_t)\tilde{s}_t^m r_t^{R,f}\\
    &+w_{\mathrm{ref}}g_t\tilde{s}_t^l r_t^{\mathrm{ref}}\\
    &+w_{\mathrm{prog}}\tilde{s}_t^l
    (0.25+0.75g_t)r_t^{\mathrm{prog}}.
\end{aligned}
\label{eq:task_reward}
\end{equation}
The complete reward adds the explicit safety reward
and regularization terms:
\begin{equation}
\begin{aligned}
    r_t
    ={}&r_t^{\mathrm{task}}
    +w_s\left[(1-g_t)s_t^m+g_ts_t^l\right]\\
    &+\sum_i w_i r_{i,t}^{\mathrm{reg}}.
\end{aligned}
\label{eq:total_reward}
\end{equation}
Here, the $w$ terms are reward weights.
The regularization terms penalize excessive torque,
power, action variation, foot slip, non-flat support,
insufficient leg separation, joint-limit violations,
and unsafe termination.

\begin{table}[t]
\centering
\caption{Reward terms and scales used for policy training.}
\label{tab:reward_terms}
\scriptsize
\setlength{\tabcolsep}{2pt}
\renewcommand{\arraystretch}{0.95}
\begin{tabular}{@{}p{0.26\columnwidth}p{0.50\columnwidth}r@{}}
\toprule
\textbf{Term} & \textbf{Definition} & \textbf{Scale} \\
\midrule
\multicolumn{3}{l}{\textit{Tracking}} \\
Position tracking
& $r_t^p=e^{-e_t^p/\sigma_p^2}+e^{-5e_t^p/\sigma_p^2}$
& $6.0$ \\
Coarse orientation
& $r_t^{R,c}=\exp[-(e_t^R/1.5)^2]$
& $6.0$ \\
Fine orientation
& $r_t^{R,f}=e^{-e_t^p/0.5}\exp[-(e_t^R/0.25)^2]$
& $6.0$ \\
\midrule
\multicolumn{3}{l}{\textit{Phase}} \\
Reference tracking
& $r_t^{\mathrm{ref}}=\exp(-e_t^{\mathrm{ref}}/\sigma_{\mathrm{ref}}^2)$
& $10.0$ \\
\midrule
\multicolumn{3}{l}{\textit{Progress}} \\
Best-so-far progress & $r_t^{\mathrm{prog}}$ in Eq.~\eqref{eq:progress_reward} & $15.0$ \\
\midrule
\multicolumn{3}{l}{\textit{Safety}} \\
Phase safety & $(1-g_t)s_t^m+g_ts_t^l$ in Eq.~\eqref{eq:total_reward} & $2.0$ \\
\midrule
\multicolumn{3}{l}{\textit{Regularization}} \\
Joint torque & $\lVert\boldsymbol{\tau}_t\rVert_2^2$ & $-1.0\times10^{-5}$ \\
Joint power & $\sum_j |\tau_{t,j}\dot q_{t,j}|$ & $-2.0\times10^{-5}$ \\
Action variation & $\lVert\mathbf{a}_t-\mathbf{a}_{t-1}\rVert_2^2$ & $-2.0\times10^{-2}$ \\
Foot slip & $\sum_{f\in\mathcal{F}_c}\lVert\mathbf{v}_{f,xy}\rVert_2^2$ & $-0.5$ \\
Non-flat support & $\sum_{f\in\mathcal{F}_c}\lVert\mathbf{z}^{w}_{f,xy}\rVert_2^2$ & $-1.0$ \\
Leg separation & $[d_{\min}-|y_t^L-y_t^R|]_+$ & $-10.0$ \\
Joint limits & $\sum_j([q_{t,j}-q_j^{\max}]_+ + [q_j^{\min}-q_{t,j}]_+)$ & $-1.0$ \\
Unsafe termination & $\mathbf{1}_{\mathrm{terminated}}$ & $-100$ \\
\bottomrule
\end{tabular}
\end{table}

\section{Experiments and results}
\label{sec:experiments}

We evaluate our controller in both simulation and real-world deployment.
In simulation, we conduct ablation studies to investigate the effects
of reward-gated training and temporal context estimation on
end-effector tracking and whole-body stability.
We then deploy the learned policy on the real bipedal manipulator and
demonstrate its ability to execute diverse manipulation tasks under
different high-level command sources, including VR teleoperation,
learned diffusion-policy commands, and scripted end-effector
trajectories.

\subsection{Experimental Setup}

\begin{figure}[t]
    \centering
    \includegraphics[width=0.9\columnwidth]{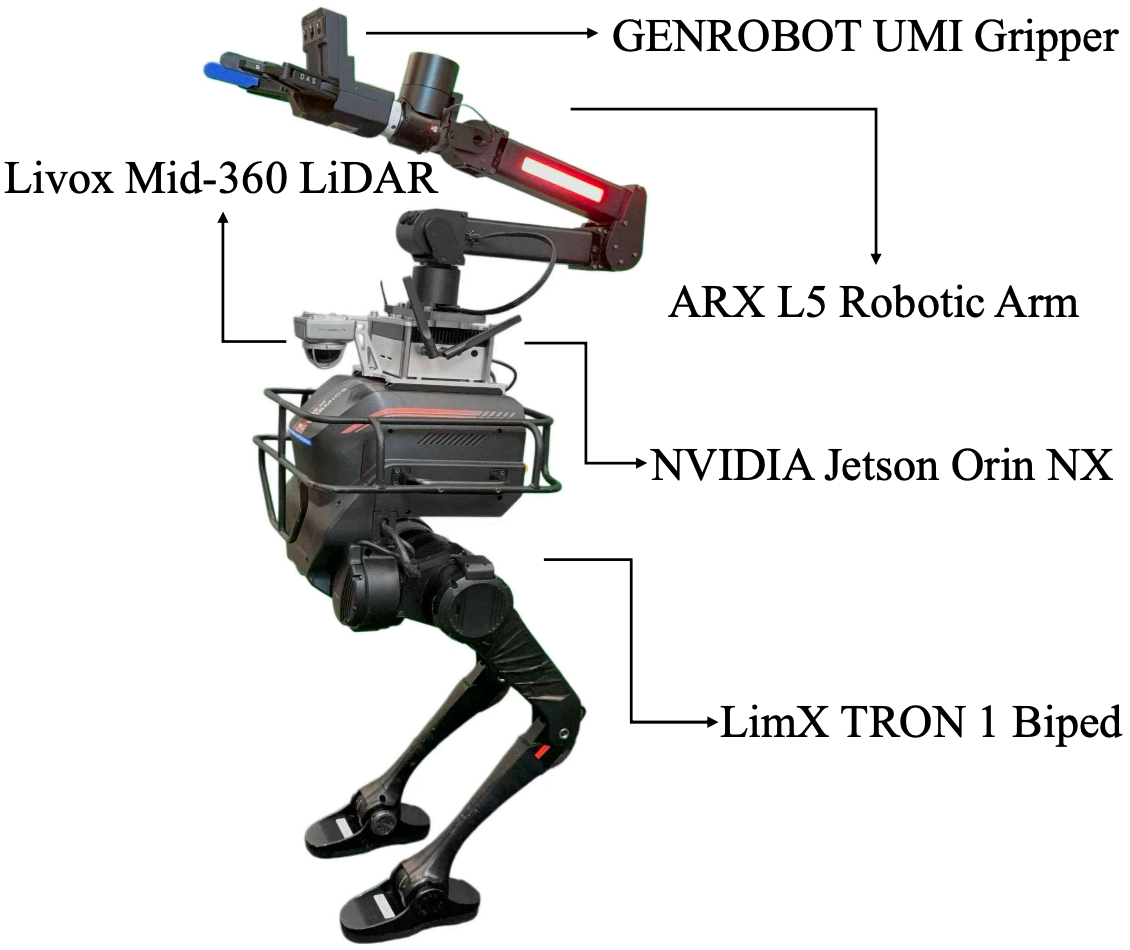}
    \caption{Real-world platform consisting of a LimX TRON 1 biped,
    an ARX L5 robotic arm, a GENROBOT UMI gripper, a Livox Mid-360
    LiDAR, and an NVIDIA Jetson Orin NX for onboard state estimation
    and whole-body policy inference.}
    \label{fig:setup}
\end{figure}

Our real-world platform, shown in Fig.~\ref{fig:setup}, consists of a
LimX TRON 1 biped equipped with an ARX L5 robotic arm, a GENROBOT UMI
gripper~\cite{chi2024umi}, a Livox Mid-360 LiDAR, and an NVIDIA Jetson
Orin NX. The Jetson runs onboard state estimation and the learned
whole-body policy, which jointly controls the biped and robotic arm.
FAST-LIO2~\cite{xu2022fastlio2} provides real-time LiDAR--inertial
localization, while LiDAR measurements also provide ground-height estimation. The estimated
base pose and arm joint states are combined through forward kinematics to recover the
end-effector pose in the world frame for tracking the commanded 6-DoF
end-effector target.

We train our controller in Isaac Lab~\cite{mittal2025isaaclab} using
PPO~\cite{schulman2017ppo} with 8,192 parallel environments for
20,000 policy iterations. Training on a single NVIDIA GeForce RTX 5090
GPU takes approximately 28 hours. The policy runs at 50~Hz, while the
resulting joint-position commands are tracked at 500~Hz by the
low-level PD controller.

\begin{table*}[t]
\centering
\caption{Simulation evaluation and ablation results.}
\label{tab:simulation_ablation}
\footnotesize
\setlength{\tabcolsep}{4pt}
\renewcommand{\arraystretch}{1.08}
\begin{tabular}{llcccccc}
\toprule
\textbf{Study}
& \textbf{Method}
& \textbf{Success} $\uparrow$
& \textbf{Pos. Mean} $\downarrow$
& \textbf{Pos. P95} $\downarrow$
& \textbf{Ori. Mean} $\downarrow$
& \textbf{Ori. P95} $\downarrow$
& \textbf{Action Variation} $\downarrow$ \\
&
&
\textbf{(\%)}
& \textbf{(cm)}
& \textbf{(cm)}
& \textbf{(deg)}
& \textbf{(deg)}
& \\
\midrule

Reward
& Matched additive
& 82.73 $\pm$ 0.93
& 3.23 $\pm$ 0.02
& 14.35 $\pm$ 0.25
& \textbf{2.12 $\pm$ 0.12}
& \textbf{5.08 $\pm$ 0.59}
& 0.218 $\pm$ 0.010 \\

& Controller (reward gating)
& \textbf{88.30 $\pm$ 2.50}
& \textbf{2.85 $\pm$ 0.10}
& \textbf{5.38 $\pm$ 0.35}
& 3.19 $\pm$ 0.04
& 6.49 $\pm$ 0.25
& \textbf{0.165 $\pm$ 0.006} \\

\midrule

Temporal context
& No latent
& 69.87 $\pm$ 1.19
& 3.65 $\pm$ 0.07
& 7.51 $\pm$ 0.07
& 3.85 $\pm$ 0.08
& 7.58 $\pm$ 0.46
& 0.236 $\pm$ 0.004 \\

& DreamWaQ/CENet
& 68.90 $\pm$ 2.16
& 3.83 $\pm$ 0.14
& 8.14 $\pm$ 0.58
& 4.22 $\pm$ 0.17
& 9.06 $\pm$ 0.76
& 0.169 $\pm$ 0.007 \\

& GRU only
& 73.23 $\pm$ 3.72
& 3.59 $\pm$ 0.13
& 6.90 $\pm$ 0.28
& 3.42 $\pm$ 0.12
& 7.66 $\pm$ 0.40
& 0.173 $\pm$ 0.008 \\

& Transformer only
& 84.43 $\pm$ 1.66
& 3.21 $\pm$ 0.14
& 6.33 $\pm$ 0.36
& \textbf{2.77 $\pm$ 0.04}
& \textbf{5.23 $\pm$ 0.13}
& \textbf{0.160 $\pm$ 0.006} \\

& Controller (Transformer+GRU)
& \textbf{88.30 $\pm$ 2.50}
& \textbf{2.85 $\pm$ 0.10}
& \textbf{5.38 $\pm$ 0.35}
& 3.19 $\pm$ 0.04
& 6.49 $\pm$ 0.25
& 0.165 $\pm$ 0.006 \\

\midrule

Upper bound
& Privileged oracle
& 94.53 $\pm$ 1.54
& 2.51 $\pm$ 0.10
& 4.78 $\pm$ 0.24
& 3.14 $\pm$ 0.12
& 5.91 $\pm$ 0.36
& 0.112 $\pm$ 0.009 \\

\bottomrule
\end{tabular}
\end{table*}

\begin{figure*}[!t]
    \begin{minipage}[t]{0.48\textwidth}
        \centering
        \includegraphics[width=0.98\linewidth,height=0.145\textheight,
        keepaspectratio]{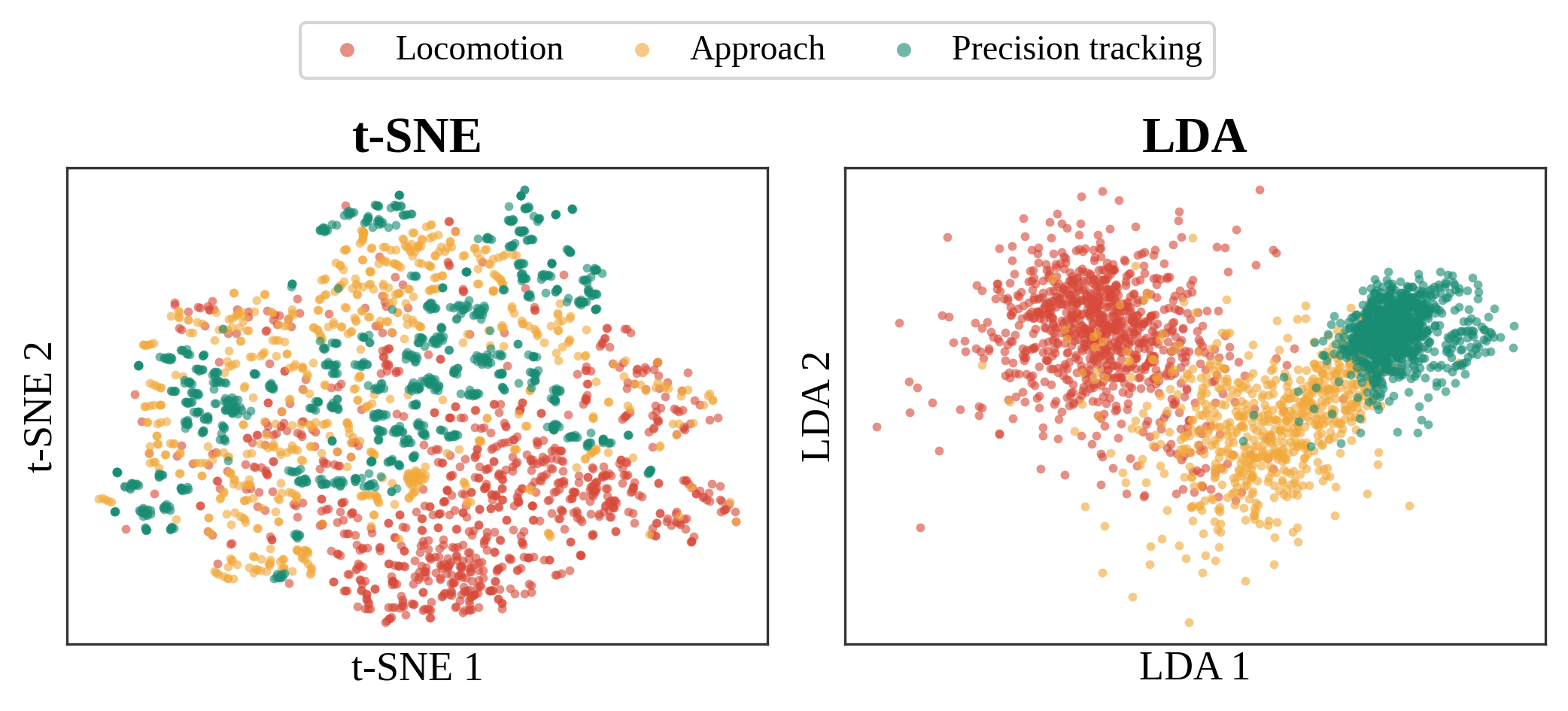}
        \captionof{figure}{Visualization of the learned temporal latent
        representation. Left: an unsupervised t-SNE projection of the
        latent features. Right: an LDA projection using task phases
        defined according to end-effector position error.}
        \label{fig:latent_visualization}
    \end{minipage}
    \hfill
    \begin{minipage}[t]{0.48\textwidth}
        \centering
        \includegraphics[width=0.90\linewidth,height=0.145\textheight,
        keepaspectratio]{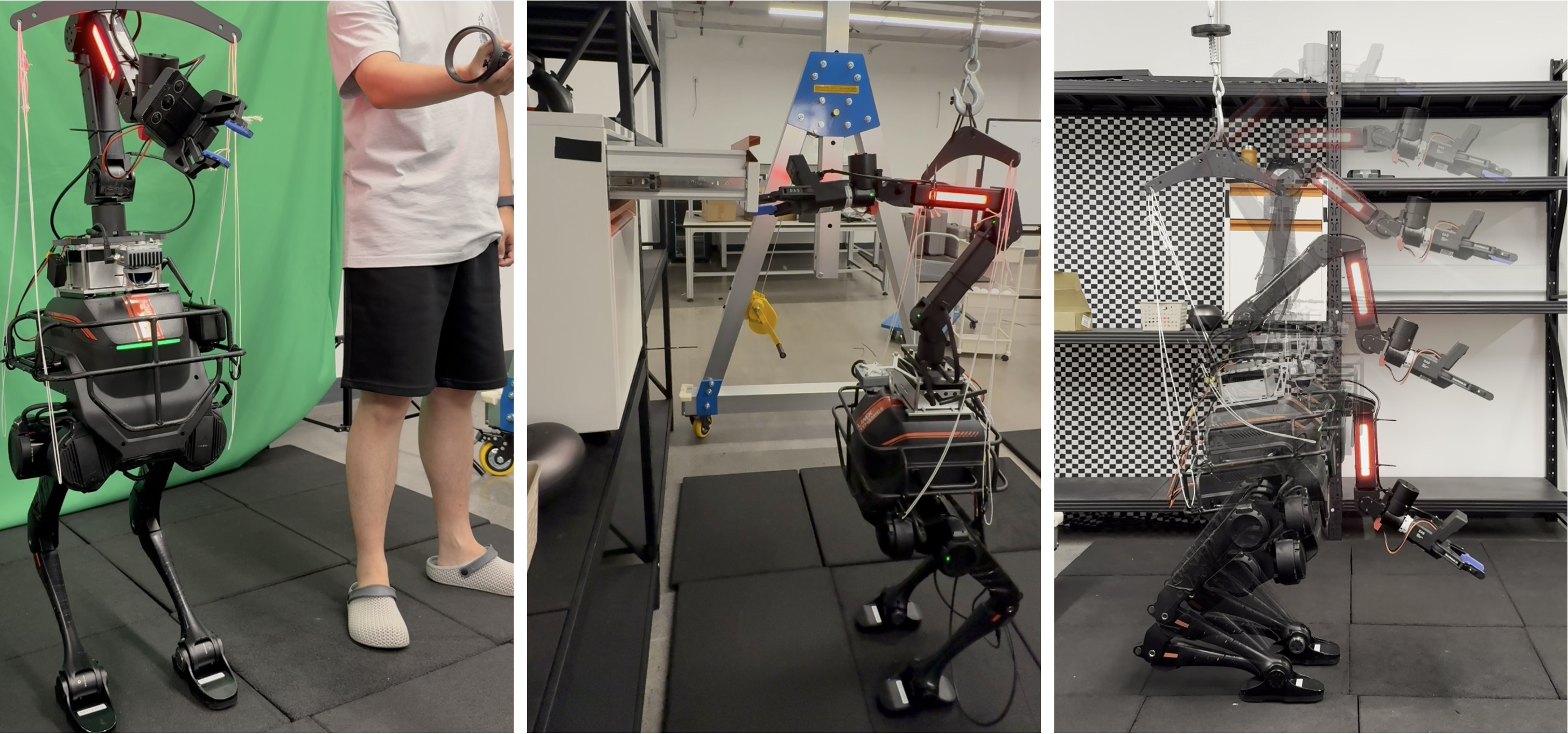}
        \captionof{figure}{Unified controller interface for real-world
        deployment. From left to right: Quest teleoperation, UMI-based
        diffusion policy, and scripted trajectories tracking.}
        \label{fig:interfaces}
    \end{minipage}

    \vspace{6pt}
    \centering
    \includegraphics[width=\textwidth,trim=0 13bp 0 0,clip]
    {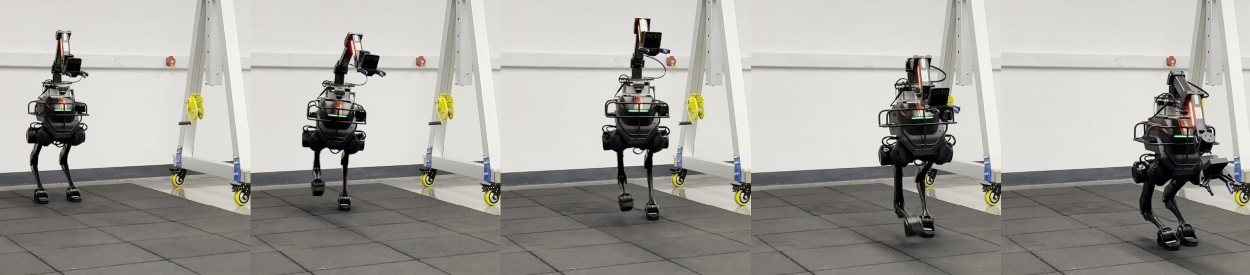}
    \captionof{figure}{Successive frames of real-world whole-body
    adaptation during end-effector tracking. The controller autonomously
    coordinates postural adjustment and stepping using only the commanded
    6-DoF end-effector pose.}
    \label{fig:locomotion_sequence}
\end{figure*}

\subsection{Evaluation in Simulation}
\label{sec:simulation}

We evaluate the policy trained with our method in simulation to quantify task-level tracking
performance, control smoothness, and the contribution of the two
key design choices: reward gating and temporal context estimation.
All methods are evaluated under the same target distribution,
domain-randomization settings, episode duration, and success
criteria.

\textbf{Evaluation protocol and metrics.}
For each method, we train three policies using independent random seeds and evaluate the final checkpoint after 20,000 training iterations. Each policy is evaluated over 1,000 episodes under the same target distribution and domain-randomization settings, resulting in 3,000 evaluation episodes for each method. The results are reported as mean $\pm$ standard deviation over the three training seeds.
We report task success rate together with the mean and
95th-percentile (P95) end-effector tracking errors.
An episode is considered successful if the position and orientation
errors remain below $5$ cm and $7^\circ$, respectively, during the
final $1$ s of the rollout.
Position error is measured as the Euclidean distance, while
orientation error is computed as the geodesic distance between
rotation matrices. At each step, action variation is computed as
$\|\mathbf{a}_t-\mathbf{a}_{t-1}\|_2^2$ and averaged over the
evaluation rollouts; lower values indicate smoother control.

\textbf{Reward-gating ablation.}
We first compare the proposed reward-gating formulation with a
matched additive reward containing the same reward terms and
coefficients. As shown in Table~\ref{tab:simulation_ablation},
reward gating increases the success rate from $82.73\%$ to
$88.30\%$ and reduces the mean position error from $3.23$ cm to
$2.85$ cm. The improvement is particularly pronounced in the
tail of the error distribution: position P95 decreases from
$14.35$ cm to $5.38$ cm. At the same time, action variation is
reduced from $0.218$ to $0.165$, indicating
smoother policy outputs.

Interestingly, the matched additive reward achieves lower
orientation error than reward gating. This result highlights a
trade-off between minimizing individual tracking terms and
maximizing task-level whole-body performance. The additive
objective can achieve accurate orientation tracking on successful
rollouts, yet exhibits lower overall success and substantially
larger tail position errors. In contrast, reward gating prioritizes
coordinated whole-body adaptation when multiple objectives
compete, resulting in higher task completion and fewer large
tracking failures.

\textbf{Temporal-context ablation.}
We next evaluate the contribution of temporal context using
no-latent, CENet~\cite{dreamwaq}, GRU-only, Transformer-only,
and the proposed Transformer--GRU estimator. Removing temporal
context reduces the success rate to $69.87\%$ and increases the
action variation to $0.236$. Introducing either recurrent or
attention-based temporal encoding improves performance, showing
that historical observations are important for inferring the
latent state of the whole-body system.

The full Transformer--GRU estimator achieves the highest
non-oracle success rate of $88.30\%$, together with the lowest
mean and P95 position errors of $2.85$ cm and $5.38$ cm,
respectively. Compared with the Transformer-only variant, the
full estimator improves success by $3.87$ percentage points and
substantially reduces both mean and tail position errors, although
the Transformer-only model attains slightly lower orientation
error and action variation. These results suggest that attention
over a finite observation window and recurrent memory provide
complementary temporal information: the Transformer captures
structured short-term dependencies, while the GRU maintains
information beyond the fixed observation window.

Finally, the privileged oracle provides ground-truth simulation
information to the actor and reaches a $94.53\%$ success rate,
serving as an approximate upper bound. The relatively small gap
in task success between our controller and the oracle suggests that the
proposed temporal estimator recovers a substantial portion of the
information required for whole-body adaptation from observation
history alone.

\textbf{Latent representation analysis.}
To examine the information encoded by the temporal context estimator,
we visualize its latent features using t-SNE and linear discriminant
analysis (LDA), as shown in Fig.~\ref{fig:latent_visualization}.
The samples are categorized into locomotion, approach, and precision
tracking phases according to the current end-effector position error.
The unsupervised t-SNE projection reveals locally structured and
partially overlapping regions, whereas the supervised LDA projection
more clearly exposes the task-phase-dependent organization of the
representation. Importantly, the overlap between adjacent phases
suggests a continuous transition rather than three completely
independent motion classes.

\subsection{Real-World Demonstrations}
\label{sec:real_world}

We evaluate our method on the real robot from two complementary perspectives:
(i) whether it can provide a unified task-space interface for heterogeneous
high-level command sources, and (ii) whether the learned whole-body behavior
provides a larger and more stable manipulation workspace than a conventional
floating-base + inverse-kinematics (IK) controller.

\textbf{Unified task-space interface.}
As illustrated in Fig.~\ref{fig:interfaces}, we drive the same controller
using three distinct command sources: Quest-based human teleoperation, a
diffusion policy trained from UMI-style demonstrations~\cite{chi2024umi,chi2023diffusionpolicy},
and scripted trajectories. Despite their different origins, commands from all high-level sources are transformed into the instantaneous base frame before being passed to our controller, ensuring a common task-space representation independent of the command source. No source-specific base-velocity,
gait, footstep, or whole-body joint commands are required, and the controller is used without modification across all three interfaces.

Across these interfaces, we demonstrate a diverse set of real-world
manipulation tasks. With Quest-based teleoperation, the robot picks up a
plush toy from the ground, takes a step, and places it onto a shelf,
requiring coordinated squatting, reaching, locomotion, and recovery. We
also teleoperate the robot to pick up a whiteboard eraser and wipe a blackboard,
demonstrating continuous task-space motion while maintaining whole-body
balance. With the UMI-based diffusion policy, the robot autonomously
pushes a cabinet door closed using only the predicted end-effector
commands. Finally, scripted trajectories command the end effector to
follow repeated upward and downward motions, allowing us to evaluate
whole-body adaptation to targets spanning different heights. Despite the
different command sources and task requirements, all behaviors are
executed by the same controller through the same 6-DoF end-effector
interface. Figure~\ref{fig:locomotion_sequence} shows successive frames
of the controller autonomously recruiting lower-body motion and stepping
while tracking an end-effector target. Additional real-world
demonstrations are provided in the supplementary material.

\begin{figure}[t]
    \centering
    \includegraphics[width=0.90\columnwidth]{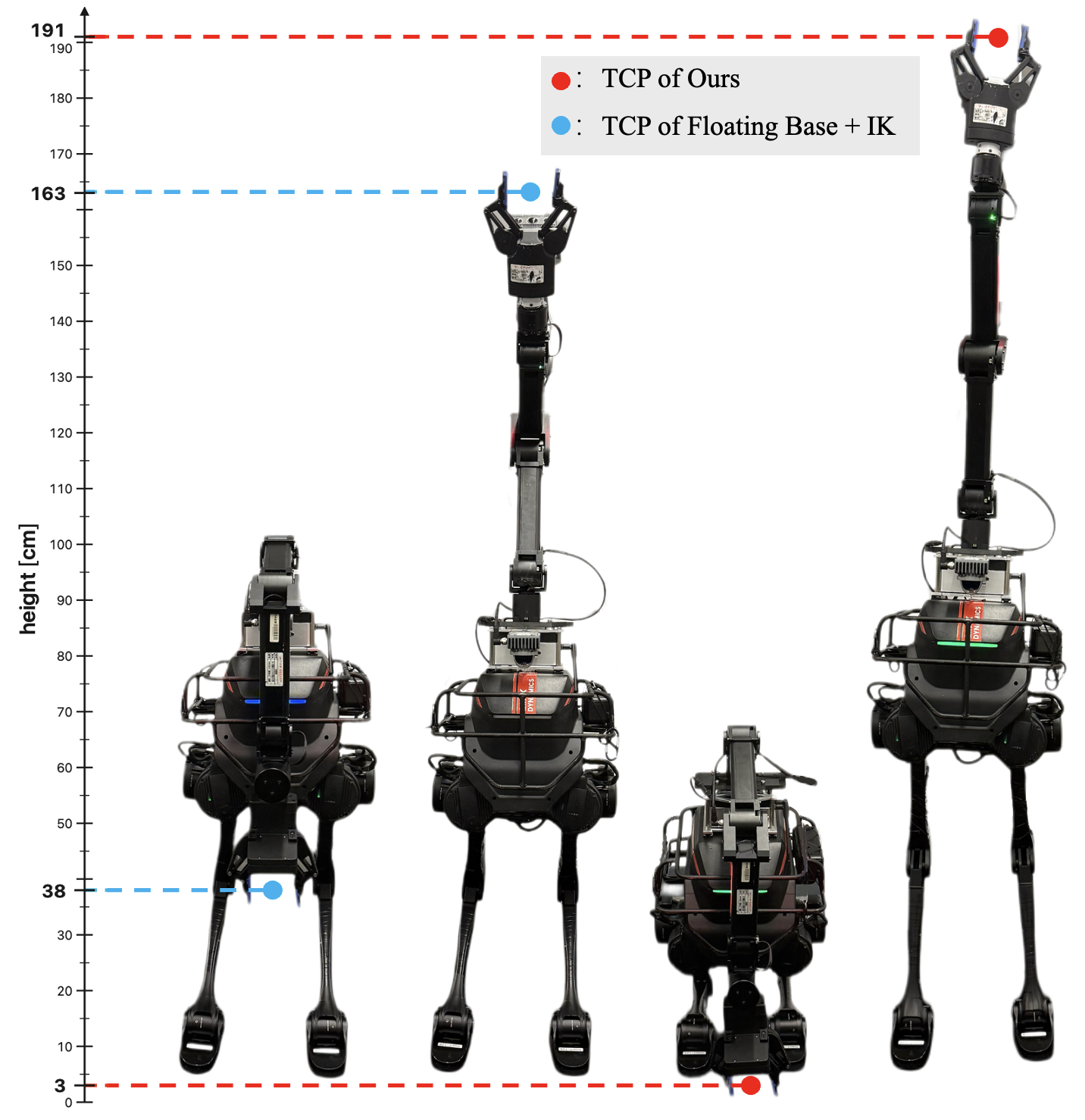}
    \caption{Real-world comparison of vertical end-effector reachability
    between our controller and a floating-base + IK baseline. The controller exploits autonomous
    whole-body motion to extend the reachable height from $3$ to
    $191\,\mathrm{cm}$, compared with $38$ to $163\,\mathrm{cm}$ for the
    baseline.}
    \label{fig:workspace}
\end{figure}

\textbf{Whole-body workspace and stability.}
We further compare our controller with a floating-base + IK baseline to evaluate the
benefit of learned whole-body coordination for task-space reachability.
As shown in Fig.~\ref{fig:workspace}, the floating-base + IK controller
achieves a vertical end-effector workspace of approximately
$38$--$163\,\mathrm{cm}$. In comparison, our controller extends the reachable range
to approximately $3$--$191\,\mathrm{cm}$ by autonomously coordinating
squatting, standing extension, and arm motion. This substantially
larger workspace is achieved without explicitly commanding the base posture
or locomotion.
Beyond reachability, we observe that the floating-base + IK baseline
exhibits pronounced oscillatory motion near challenging workspace
boundaries, whereas our controller maintains substantially smoother and more stable
whole-body behavior. This comparison highlights that our controller does not merely
increase geometric reachability, but learns to exploit whole-body mobility
while preserving balance and coordinated end-effector motion.

\section{Conclusion}

We presented an end-effector-conditioned whole-body controller trained
with reinforcement learning for bipedal mobile manipulation that directly maps a
single 6-DoF end-effector target to coordinated whole-body actions without
explicit base-velocity, gait, or footstep commands. The controller combines
reward-gated training with a Transformer--GRU temporal context estimator
to support accurate manipulation while maintaining dynamic balance.
Experiments in simulation and on a real bipedal manipulator demonstrate
that the same controller can autonomously produce arm-dominant reaching,
postural adaptation, squatting, and stepping, while also serving as a
common task-space interface for human, learned, and scripted high-level
commands. Future work will extend the framework toward more dynamic loco-manipulation behaviors and improve task-space tracking precision for fine-grained manipulation, particularly when following learned high-level commands.








\bibliographystyle{IEEEtran}
\bibliography{reference}

@misc{sun2025ulc,
  title={ULC: A Unified and Fine-Grained Controller for Humanoid Loco-Manipulation},
  author={Wandong Sun and Luying Feng and Baoshi Cao and Yang Liu and Yaochu Jin and Zongwu Xie},
  year={2025},
  eprint={2507.06905},
  archivePrefix={arXiv},
  primaryClass={cs.RO},
  url={https://arxiv.org/abs/2507.06905}
}

@article{cao2026hiwet,
  title={HiWET: Hierarchical World-Frame End-Effector Tracking for Long-Horizon Humanoid Loco-Manipulation},
  author={Cao, Zhanxiang and Yan, Liyun and Zhang, Yang and Chen, Sirui and Ma, Jianming and Zhan, Tianyue and Fu, Shengcheng and Jia, Yufei and Lu, Cewu and Gao, Yue},
  journal={arXiv preprint arXiv:2602.06341},
  year={2026}
}

@inproceedings{fu2023deep,
  title={Deep Whole-Body Control: Learning a Unified Policy for Manipulation and Locomotion},
  author={Fu, Zipeng and Cheng, Xuxin and Pathak, Deepak},
  booktitle={Proceedings of The 6th Conference on Robot Learning},
  pages={138--149},
  year={2023},
  volume={205},
  series={Proceedings of Machine Learning Research},
  publisher={PMLR}
}

@inproceedings{liu2025visualwbc,
  title={Visual Whole-Body Control for Legged Loco-Manipulation},
  author={Liu, Minghuan and Chen, Zixuan and Cheng, Xuxin and Ji, Yandong and Qiu, Ri-Zhao and Yang, Ruihan and Wang, Xiaolong},
  booktitle={Proceedings of The 8th Conference on Robot Learning},
  pages={234--257},
  year={2025},
  volume={270},
  series={Proceedings of Machine Learning Research},
  publisher={PMLR}
}

@inproceedings{portela2025whole,
  title={Whole-Body End-Effector Pose Tracking},
  author={Portela, Tifanny and Cramariuc, Andrei and Mittal, Mayank and Hutter, Marco},
  booktitle={2025 IEEE International Conference on Robotics and Automation (ICRA)},
  pages={11205--11211},
  year={2025},
  organization={IEEE},
  doi={10.1109/ICRA55743.2025.11127964}
}

@inproceedings{ha2025umionlegs,
  title={UMI-on-Legs: Making Manipulation Policies Mobile with Manipulation-Centric Whole-body Controllers},
  author={Ha, Huy and Gao, Yihuai and Fu, Zipeng and Tan, Jie and Song, Shuran},
  booktitle={Proceedings of The 8th Conference on Robot Learning},
  pages={5254--5270},
  year={2025},
  volume={270},
  series={Proceedings of Machine Learning Research},
  publisher={PMLR}
}

@article{jiang2025rfm,
  title={Learning Whole-Body Loco-Manipulation for Omni-Directional Task Space Pose Tracking With a Wheeled-Quadrupedal-Manipulator},
  author={Jiang, Kaiwen and Fu, Zhen and Guo, Junde and Zhang, Wei and Chen, Hua},
  journal={IEEE Robotics and Automation Letters},
  volume={10},
  number={2},
  pages={1481--1488},
  year={2025},
  doi={10.1109/LRA.2024.3519856}
}

@misc{jang2025seec,
  title={SEEC: Stable End-Effector Control with Model-Enhanced Residual Learning for Humanoid Loco-Manipulation},
  author={Jang, Jaehwi and Wang, Zhuoheng and Zhou, Ziyi and Wu, Feiyang and Zhao, Ye},
  year={2025},
  eprint={2509.21231},
  archivePrefix={arXiv},
  primaryClass={cs.RO}
}

@misc{luo2026ceer,
  title={CEER: Compliant End-Effector and Root Control as a Unified Interface for Hierarchical Humanoid Loco-Manipulation},
  author={Luo, Xinyuan and Chen, Xingrui and Yin, Xunjian and Wu, Hongxuan and Xia, Boxi and Chen, Zhuoqun and Li, Jinzhou and Chen, Boyuan and Cheng, Xianyi},
  year={2026},
  eprint={2605.19981},
  archivePrefix={arXiv},
  primaryClass={cs.RO}
}

@inproceedings{dreamwaq,
  title={DreamWaQ: Learning Robust Quadrupedal Locomotion With Implicit Terrain Imagination via Deep Reinforcement Learning},
  author={Nahrendra, I Made Aswin and Yu, Byeongho and Myung, Hyun},
  booktitle={2023 IEEE International Conference on Robotics and Automation (ICRA)},
  pages={5078--5084},
  year={2023},
  organization={IEEE},
  doi={10.1109/ICRA48891.2023.10161144}
}

@misc{schulman2017ppo,
  title={Proximal Policy Optimization Algorithms},
  author={Schulman, John and Wolski, Filip and Dhariwal, Prafulla and
          Radford, Alec and Klimov, Oleg},
  year={2017},
  eprint={1707.06347},
  archivePrefix={arXiv},
  primaryClass={cs.LG}
}

@inproceedings{pinto2018asymmetric,
  title={Asymmetric Actor Critic for Image-Based Robot Learning},
  author={Pinto, Lerrel and Andrychowicz, Marcin and Welinder, Peter and
          Zaremba, Wojciech and Abbeel, Pieter},
  booktitle={Proceedings of Robotics: Science and Systems},
  year={2018},
  address={Pittsburgh, Pennsylvania},
  doi={10.15607/RSS.2018.XIV.008}
}

@misc{mittal2025isaaclab,
  title={Isaac Lab: A GPU-Accelerated Simulation Framework for
         Multi-Modal Robot Learning},
  author={Mittal, Mayank and Roth, Pascal and Tigue, James and others},
  year={2025},
  eprint={2511.04831},
  archivePrefix={arXiv},
  primaryClass={cs.RO}
}

@inproceedings{vaswani2017attention,
  title={Attention Is All You Need},
  author={Vaswani, Ashish and Shazeer, Noam and Parmar, Niki and
          Uszkoreit, Jakob and Jones, Llion and Gomez, Aidan N. and
          Kaiser, Lukasz and Polosukhin, Illia},
  booktitle={Advances in Neural Information Processing Systems},
  volume={30},
  year={2017}
}

@inproceedings{cho2014learning,
  title={Learning Phrase Representations using RNN Encoder--Decoder
         for Statistical Machine Translation},
  author={Cho, Kyunghyun and van Merri{\"e}nboer, Bart and Gulcehre,
          Caglar and Bahdanau, Dzmitry and Bougares, Fethi and
          Schwenk, Holger and Bengio, Yoshua},
  booktitle={Proceedings of the 2014 Conference on Empirical Methods
             in Natural Language Processing (EMNLP)},
  pages={1724--1734},
  year={2014},
  address={Doha, Qatar},
  publisher={Association for Computational Linguistics},
  doi={10.3115/v1/D14-1179}
}

@inproceedings{chi2024umi,
  title={Universal Manipulation Interface: In-The-Wild Robot Teaching Without In-The-Wild Robots},
  author={Chi, Cheng and Xu, Zhenjia and Pan, Chuer and Cousineau, Eric
          and Burchfiel, Benjamin and Feng, Siyuan and Tedrake, Russ
          and Song, Shuran},
  booktitle={Proceedings of Robotics: Science and Systems},
  year={2024},
  address={Delft, Netherlands},
  doi={10.15607/RSS.2024.XX.045}
}

@inproceedings{chi2023diffusionpolicy,
  title={Diffusion Policy: Visuomotor Policy Learning via Action Diffusion},
  author={Chi, Cheng and Feng, Siyuan and Du, Yilun and Xu, Zhenjia
          and Cousineau, Eric and Burchfiel, Benjamin C. M.
          and Song, Shuran},
  booktitle={Proceedings of Robotics: Science and Systems},
  year={2023},
  address={Daegu, Republic of Korea},
  doi={10.15607/RSS.2023.XIX.026}
}

@article{xu2022fastlio2,
  title={FAST-LIO2: Fast Direct LiDAR-Inertial Odometry},
  author={Xu, Wei and Cai, Yixi and He, Dongjiao and Lin, Jiarong and Zhang, Fu},
  journal={IEEE Transactions on Robotics},
  volume={38},
  number={4},
  pages={2053--2073},
  year={2022},
  doi={10.1109/TRO.2022.3141876}
}

@inproceedings{zhou2019continuity,
  title={On the Continuity of Rotation Representations in Neural Networks},
  author={Zhou, Yi and Barnes, Connelly and Lu, Jingwan and
          Yang, Jimei and Li, Hao},
  booktitle={Proceedings of the IEEE/CVF Conference on Computer Vision
             and Pattern Recognition (CVPR)},
  pages={5745--5753},
  year={2019}
}

@inproceedings{kingma2014autoencoding,
  title     = {Auto-Encoding Variational Bayes},
  author    = {Kingma, Diederik P. and Welling, Max},
  booktitle = {International Conference on Learning Representations (ICLR)},
  year      = {2014}
}

\end{document}